\documentclass[lettersize,journal]{IEEEtran}
\usepackage{amsmath,amsfonts}
\usepackage{algorithmic}
\usepackage{array}
\usepackage[caption=false,font=normalsize,labelfont=sf,textfont=sf]{subfig}
\usepackage{textcomp}
\usepackage{stfloats}
\usepackage{url}
\usepackage{verbatim}
\usepackage{graphicx}

\usepackage{cite}
\usepackage{multirow}
\usepackage{makecell}
\usepackage{color}
\usepackage{algorithm}
\usepackage{hyperref}
\usepackage{booktabs}
\usepackage{graphicx}
\usepackage{wrapfig}
\usepackage{orcidlink}

\def\BibTeX{{\rm B\kern-.05em{\sc i\kern-.025em b}\kern-.08em
    T\kern-.1667em\lower.7ex\hbox{E}\kern-.125emX}}
\usepackage{balance}
\begin{document}
\title{$\mathbf{S^2LM}$: Towards Semantic-Payload Steganography via Large Language Models}
\author{Huanqi Wu\orcidlink{0009-0008-4518-3273}, Huangbiao Xu\orcidlink{0000-0002-3717-8713}, Yongliang Xu\orcidlink{0000-0001-7571-5498}, Jiaxin Cai\orcidlink{0009-0003-5697-1746}, \\Xiao Ke\orcidlink{0000-0001-9059-5391},~\IEEEmembership{Senior Member,~IEEE}, and Hang Cheng\orcidlink{0000-0003-2443-2820},~\IEEEmembership{Member,~IEEE}
\thanks{This work was supported in part by the National Key Research and Development Plan of China under Grant 2021YFB3600503, in part by the National Natural Science Foundation of China under Grant 61972097 and U21A20472, in part by the Major Scientific Research Project for Technology Promotes Police under Grant 2025YZ040003, 2024YZ040001, in part by the Natural Science Foundation of Fujian Province under Grant 2025J01536.}
\thanks{Huanqi Wu, Huangbiao Xu, Jiaxin Cai, Xiao Ke are with the College of Computer and Data Science, Fuzhou University, Fuzhou 350116, China. Huanqi Wu, Huangbiao Xu, Xiao Ke are also with the Fujian Provincial Key Laboratory of Networking Computing and Intelligent Information Processing, the Engineering Research Center of Big Data Intelligence, Ministry of Education, Fuzhou 350116, China. (e-mail: wuhuanqi135@gmail.com; huangbiaoxu.chn@gmail.com; jiaxincai528@163.com; kex@fzu.edu.cn)}
\thanks{Yongliang Xu and Hang Cheng are with the School of Mathematics and Statistics, Fuzhou University, Fuzhou 350108, China (e-mail: 230310014@fzu.edu.cn; hcheng@fzu.edu.cn).}
\thanks{Xiao Ke is the corresponding author.}
}

\markboth{Journal of \LaTeX\ Class Files,~Vol.~18, No.~9, September~2020}%
{How to Use the IEEEtran \LaTeX \ Templates}

\maketitle

\begin{figure*}[htb]
    \includegraphics[width=0.99\textwidth]{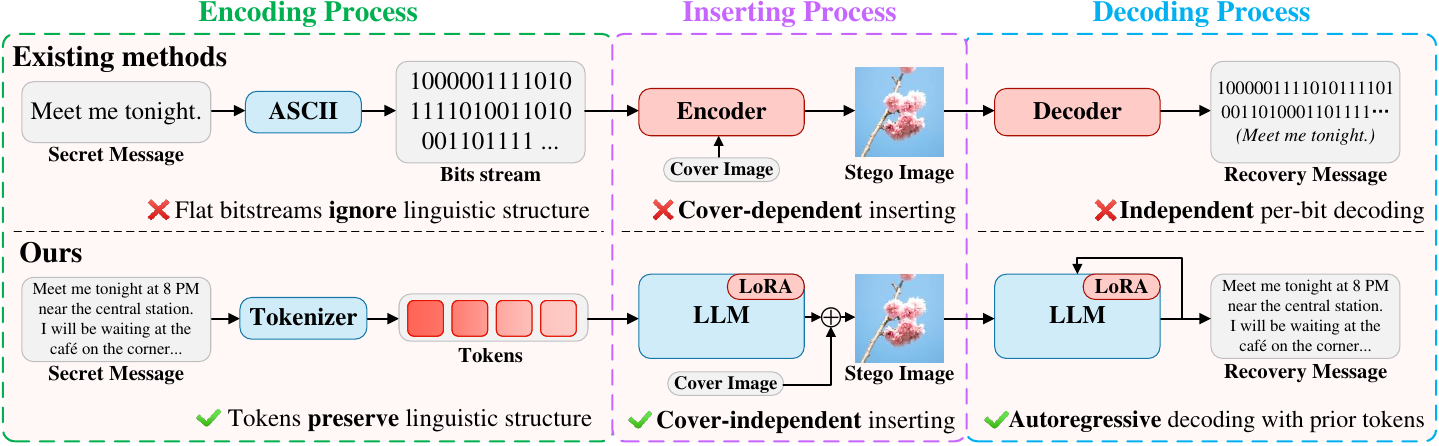}
    \caption{Overview of our $\mathrm{S^2LM}$ framework vs. existing frameworks. We divide the pipeline in three processes and show the difference between the existing methods and our $\mathrm{S^2LM}$ framework.}
    \label{fig:figure1}
\end{figure*}

\begin{abstract}
Despite remarkable progress in steganography, embedding sentence-level information into cover media remains challenging. In this work, we present the concept of \textbf{Semantic-Payload Steganography}, which aims to hide semantically meaningful and structured content, such as sentences, in cover media. Based on this concept, we present Sentence-to-Image Steganography as an instance that enables the hiding of sentence-level messages within a cover image. To accomplish this feat, we propose \textbf{S}emantic-Payload \textbf{S}teganographic \textbf{L}anguage \textbf{M}odel ($\mathbf{S^2LM}$), which leverages large language models to embed high-level textual information into images. $\mathbf{S^2LM}$ redesigns the entire pipeline, involving the LLM throughout the process to enable the hiding and recovery of sentences. To evaluate semantic-payload steganography methods, we establish a benchmark named \textbf{I}n\textbf{v}isible \textbf{T}ext (IVT), comprising a diverse set of corpus as secret messages. Experimental results demonstrate that $\mathbf{S^2LM}$ effectively recovers natural-language messages while maintaining high visual quality. The source code will be released at \url{https://github.com/aoli-gei/S2LM}.
\end{abstract}


\section{Introduction}
\label{sec:intro}

\IEEEPARstart{S}{teganography} is the art of hiding information in plain sight. Historically, it was used as a means of covert communication during wartime. Nowadays, these techniques have found broader applications in copyright protection \cite{zhang2024gs,tang2025simufreemark}, tamper detection \cite{Zhang_2024_CVPR,zhang2025omniguard}, and secure communication \cite{yu2023cross}.

With the advent of deep learning, steganography has achieved remarkable progress in recent years. Most existing methods formulate steganography as a signal-level embedding problem, where the secret message is first converted into a binary bitstream and then embedded into the carrier by an encoder \cite{wengrowski2019light,tancik2020stegastamp,zhu2018hidden}. By treating the secret message as a sequence of bits, these methods operate on low-level signal representations, where the hidden content is encoded in a structure-agnostic form, making it difficult to compactly express rich information. As a result, existing methods not only exhibit limited communication efficiency, but also fail to align with the way humans naturally convey and interpret information.

For further analysis, we take image steganography as a case study and divide the steganographic pipeline into three processes, as illustrated in Fig.~\ref{fig:figure1}. There are three significant limitations that hinder existing methods from embedding larger amounts of information into images:
1) \textcolor{black}{\textbf{Structure-agnostic encoding.}} In the encoding process, messages are converted into bitstreams via ASCII or UTF encoding. This method neglects the semantic relationships between words and is unsuitable for encoding large amounts of information, such as sentences. Natural language inherently contains rich linguistic structures such as contextual dependencies and syntactic relationships. However, after binary serialization, the original message is represented as flat bitstreams, causing these linguistic structures to be completely ignored during encoding.
2) \textcolor{black}{\textbf{Cover-dependent insertion.}} In the insertion process, the encoder takes the secret message and cover image as input to generate a stego image. However, bitstreams and images are inherently different in their representations and feature distributions. Directly performing cross-modal fusion between these two heterogeneous modalities within a single encoder may lead to suboptimal embedding performance. Moreover, as the secret payload increases, the encoder is required to embed a larger amount of information into an image with the same resolution, which further increases the difficulty of embedding.
3) \textcolor{black}{\textbf{Context-independent decoding.}} In the decoding process, existing methods typically formulate secret message recovery as a series of bit classification problems, where the decoder independently predicts the secret bit at each position based on features extracted from the stego image. Such independent prediction makes it difficult to model the dependencies among secret bits and prevents the decoder from exploiting contextual information to facilitate message recovery. Moreover, as the payload increases, the number of bits to be decoded grows accordingly, making accurate recovery of secret messages increasingly challenging. These practical limitations call for a rethinking of steganography: \textbf{\textit{must steganographic frameworks be confined to hiding low-level bitstreams?}}

To answer this question, we revisit the fundamental goal of steganography: covert communication. To enable more efficient secret communication, we propose to use \textbf{\textit{natural language}}, rather than binary bits, as the information carrier. This design not only be consistent with human communication and but also preserves the inherent linguistic structures and contextual dependencies of textual messages. Based on this perspective, we introduce \textbf{Semantic-Payload Steganography}, which aims to embed natural language messages as secret information into a carrier. Building on this idea, we formalize the \textbf{Sentence-to-Image Steganography} task as an instance of semantic-payload steganography. This task requires hiding sentence-level or even paragraph-level information in a cover image. Notably, the framework supports diverse textual content across a range of message lengths. To accomplish this task, the model must possess two key capabilities: effectively representing the linguistic structure of natural language messages and handling two fundamentally different modalities.

Recent advancements in large language models (LLMs) have spurred research on integrating them with vision backbones, leading to the rise of vision-language models (VLMs) \cite{zhang2024vision,ye2024mplug,li2022mplug,liu2024improved,liu2023visual}. These studies demonstrate that LLMs exhibit strong capabilities in semantic understanding and cross-modal reasoning, which precisely align with the requirements of semantic-payload steganography. These properties suggest that LLMs can provide a natural foundation for developing a universal steganographic framework for semantic payloads.

Building on this insight, we propose the \textbf{S}emantic-Payload \textbf{S}teganographic \textbf{L}anguage \textbf{M}odel (\textbf{$\mathbf{S^2LM}$}), a novel framework that leverages the powerful language understanding and contextual modeling capabilities of LLMs for steganography. To address the three challenges mentioned above, \textbf{$\mathrm{S^2LM}$} completely redesigns the entire pipeline of existing methods, as illustrated in Fig.~\ref{fig:figure1}. 
1) \textcolor{black}{\textbf{Token-based encoding.}} In the encoding process, we first transform the secret textual information into a sequence of discrete tokens using a tokenizer rather than a flat bitstream. This process not only captures the semantic content of the sentence but also preserves the relationships between words, enabling more effective and structured representation of the secret message.
2) \textcolor{black}{\textbf{Cover-independent insertion.}} During the insertion process, we leverage LLM to generate secret embeddings without relying on any prior of the cover image. By decoupling secret embeddings from the carrier, $\mathrm{S^2LM}$ prevents potential interference between the hidden message and the cover image.
3) \textcolor{black}{\textbf{Autoregressive decoding.}} In the decoding process, we employ the same LLM to autoregressively recover the secret message from the stego image. At each decoding step, the self-attention mechanism jointly leverages the secret features extracted from the stego image and the previously recovered tokens to predict the next token. In this way, the contextual dependencies within the secret message can be explicitly exploited to guide subsequent predictions, enabling the recovery of long-text secret messages.

Overall, $\mathrm{S^2LM}$ provides a unified framework for the sentence-to-image task, demonstrating strong capability in encoding and recovering semantically rich messages without compromising visual integrity, enabling the seamless embedding of textual information within images. We further evaluate the performance of different LLMs within the $\mathrm{S^2LM}$ framework and verify its versatility across various models.

Additionally, we establish a benchmark for semantic-payload steganography called \textbf{Invisible Text (IVT)}, which we use to validate the effectiveness of the sentence-to-image framework. Overall, our contributions are as follows:\begin{itemize}
    \item We propose the concept of \textbf{Semantic-Payload Steganography} and formalize \textbf{Sentence-to-Image Steganography} as an instance, redefining the traditional goal of steganography from embedding low-level data to concealing meaningful language content within cover media.
    
    \item We propose the $\mathbf{S^2LM}$, incorporating new capabilities to embed sentence-level information into images. It demonstrates the potential of LLMs in steganography and provides a new paradigm for concealing textual messages.

    \item We establish a benchmark for semantic-payload steganography named \textbf{Invisible Text (IVT)}, containing thousands of samples from diverse sources. This benchmark is essential for evaluation and encourages the community to explore the potential of semantic-payload steganography.

\end{itemize}

\section{Related Work}
\subsection{Image Steganography}
Steganography is a technique for concealing secret information within a carrier while preserving its visual appearance and avoiding detection. Traditional image steganography methods, such as the Least Significant Bit (LSB) algorithm \cite{wang2001image} and frequency-domain approaches \cite{patel2012steganography,chen2006dwt}, embed secret data by modifying the pixel values or transform coefficients of the cover image. Specifically, Chen \textit{et al.} proposed a dual-path super-resolution-based side-information estimation method for spatial image steganography \cite{chen2026side}. Guo \textit{et al.} proposed a coverless image steganography method that uses SHA-256-based hash generation and similarity-guided inverse image retrieval to improve robustness against steganalysis \cite{guo2024highly}. However, these hand-crafted methods tend to introduce statistical artifacts that can be easily detected by steganalysis, whereas deep learning-based methods achieve higher security. Recent advances in deep learning have produced more powerful steganographic techniques \cite{zhu2018hidden,wengrowski2019light,tancik2020stegastamp,li2026screening,lin2024automatic,Zeng2026UPSEC,liu2025fearless}, which can embed bit-level information in images with robustness against various manipulations. Qin \textit{et al.} proposed U-StegNet, a unified dual-stage image steganography framework that integrates generation-based and modification-based hiding with adaptive message partitioning \cite{qin2026u}. Yang \textit{et al.} proposed AutoStegaFont+, which automatically synthesizing glyph vectors for robust information hiding in documents \cite{yang2025synthesizing}. Zhang \textit{et al.} proposed a black-box image steganography method that exploits generative model volatility to derive pixel-wise modification costs and improve security \cite{zhang2023steganography}.

Moreover, methods like HiNet \cite{jing2021hinet}, ISN \cite{lu2021large}, and StegFormer \cite{ke2024stegformer} enable the concealment of an image within a cover image of the same resolution \cite{ni2025ninet}. Li \textit{et al.} proposed CIDS, a cover-independent DNN steganography framework that uses AI-generated cover images, symmetric encoding/decoding, and perturbation encryption to improve performance and prevent unauthorized extraction \cite{li2025cover}. Recently, several works have explored utilizing diffusion models for steganography \cite{Fan2026GIANT}, such as DGADM-GIS \cite{yuan2025dgadm} and Latent Watermark \cite{meng2025latent}. Zhou \textit{et al.} proposed CFGS, a conditional flow-based generative steganography method with frequency-domain secret embedding \cite{zhou2025conditional}.

Despite these advancements, existing methods primarily focus on embedding secret information at a low level (bits or images). In this paper, we present a novel steganographic task,  sentence-to-image steganography, which aims to conceal semantically meaningful content (\textit{e.g.}, sentence) into images. To this end, we propose $\mathrm{S^2LM}$, successfully embedding sentence-level textual data into cover images, significantly expanding the scope of steganography.

\subsection{Steganalysis}
Steganalysis aims to detect the presence of secret information hidden in media, which serves as an important technique for ensuring information security. With the development of steganography, steganalysis has also evolved accordingly to detect and identify increasingly sophisticated steganographic behaviors. Early steganalysis methods mainly relied on manually designed statistical features to identify the statistical anomalies introduced into cover media by steganographic embedding. Representative methods include chi-square analysis \cite{westfeld1999attacks} and RS analysis \cite{fridrich2001reliable}.

As steganographic algorithms evolved from simple LSB modification toward more sophisticated adaptive embedding strategies, the detection capability of conventional statistical methods became increasingly limited. To address this issue, researchers began to construct high-dimensional Rich Model features to characterize the subtle statistical artifacts. Fridrich \textit{et al.} proposed the Spatial Rich Model \cite{fridrich2012rich}, which models image residuals using high-pass filtering and co-occurrence statistics. Subsequently, Kodovský and Fridrich introduced the JPEG Rich Model \cite{kodovsky2012steganalysis}, which models rich statistical characteristics of DCT coefficients in JPEG images to improve the detection of JPEG-domain steganography. In addition, Denemark \textit{et al.} proposed the selection-channel-aware Rich Model \cite{denemark2014selection}, which incorporates selection-channel information, such as pixel modification probabilities, into feature extraction, further improving the detection of content-adaptive steganographic algorithms.

Despite their effectiveness in characterizing subtle statistical artifacts through high-dimensional features, Rich Model-based steganalysis methods rely heavily on manually designed feature extractors, and their feature engineering becomes increasingly complicated when dealing with sophisticated steganographic algorithms. With the rapid development of deep learning, researchers have increasingly employed deep neural networks to automatically learn discriminative representations of steganographic artifacts, thereby reducing the reliance on handcrafted features. Wei \textit{et al.} proposed UCNet \cite{wei2022universal}, which separately models residual information from different color channels to enable unified steganalysis of color images in both the spatial and JPEG domains. Subsequently, Wei \textit{et al.} combined pixel difference convolution with an enhanced Transformer to jointly capture local difference patterns and long-range dependencies, thereby further improving the detection performance of color image steganalysis \cite{wei2024color}. Fu \textit{et al.} introduced a steganalysis framework based on dual-path enhancement and fractal downsampling, which extracts and enhances weak steganographic residuals across multiple scales and structures for effective steganalysis in both the spatial and JPEG domains \cite{fu2024image}. 


\subsection{Large Language Models in Steganography}
The prosperity of LLMs has pushed the entire natural language processing community into new boundaries. The computer vision community has also recognized the tremendous success of LLMs. Consequently, lots of the latest research focuses on exploring the integrity of the LLM with the vision backbone. BLIP-2 \cite{li2023blip} introduces a frozen LLM to bootstrap vision-to-language learning and proposes a Q-Former to bridge the modality gap between language and vision. LLaVA \cite{liu2023visual} connects the vision backbone and Llama \cite{touvron2023llama}, achieving the general-purpose visual-language understanding. However, the LLM can also enhance the performance of the low-level vision backbone. LISA \cite{Lai2023LISARS} uses a novel segmentation token and unlocks the segmentation capability of the LLM. Zheng \textit{et al.} demonstrates the strong generalization capability of LLMs by freezing it to solve a range of low-level vision tasks \cite{zheng2024lm4lv}.

Moreover, several studies have explored to utilizes LLMs to steganography. Qi \textit{et al.} proposed SyncPool that groups prefix-related tokens and uses synchronized CSPRNG-based sampling to eliminate segmentation ambiguity in neural linguistic steganography \cite{qi2024provably}. Yang \textit{et al.} embed secret messages in semantic space using a reversible codebook and multi-agent LLM-based text synthesis \cite{yang2025novel}. Mathew \textit{et al.} show that LLMs can learn to hide information implicitly through reinforcement learning \cite{mathewsteganography}. Li \textit{et al.} realize black-box steganography by implanting backdoors into LLMs \cite{li2025black}. Additionally, several existing methods rely on handcrafted sampling strategies to hide information during text generation \cite{wang2025sparsamp, bai2024semantic, qu2025provably, wu2024generative, lin2024zero, jiang2025high}. 

However, these approaches depend on substantial manual design, such as hand-crafted sampling mechanisms or backdoor-style behaviors, and ultimately embed binary bitstreams rather than semantically meaningful language content. In this paper, we integrate LLM into the entire pipeline, utilizing them to generate secret embeddings directly, and enable the hiding and recovery of sentence-level messages.

\section{Semantic-Payload Steganography}
\noindent
\textbf{Motivation.}
The steganographic community has traditionally focused on hiding low-level information (\textit{e.g.}, at the bit or image level), overlooking the semantic structure of the hidden content. While this formulation provides a general representation for arbitrary data, it treats the hidden content in a structure-agnostic manner and does not explicitly account for the semantic relationships and contextual dependencies inherent in natural language. In contrast, we introduce a new direction termed \textbf{Semantic-Payload Steganography}, which refers to hiding semantically meaningful and structured content, such as sentences or paragraphs, in the carrier. This paradigm redefines what kind of information can be hidden. Rather than merely embedding arbitrary bitstreams, it enables the direct hiding and recovery of natural language messages, aligning more closely with the way humans naturally communicate and broadening the practical application scenarios of steganography.

\noindent
\textbf{Terminology Clarification.} It is important to clarify that our definition of semantic-payload steganography differs from previous works that have used similar terminology \cite{bai2024semantic,huo2024image,yang2025novel}. Prior studies primarily focused on enhancing the imperceptibility of steganography by leveraging semantic information from the cover media \cite{huo2024image}, or by employing text-to-image generative models for coverless steganography \cite{jiang2025generative}. However, the hidden information in these methods remains bit-level. In contrast, we define semantic-payload steganography as the task of embedding high-level semantic content, specifically natural language sentences, into carriers in this paper. The key distinctions between the two paradigms are summarized in Table~\ref{tab:terminology_comparison}.

\begin{table}[t]
  \centering
  \caption{Comparison of prior semantic steganography and our semantic-payload steganography.}
  \label{tab:terminology_comparison}
  \scriptsize
  \tabcolsep=4pt
  \renewcommand\arraystretch{1.1}
  \begin{tabular}{l|cc}
    \toprule
    \textbf{Aspect} & \makecell{\textbf{Semantic} \\ \textbf{Steganography (Bai \cite{bai2024semantic})}} & \makecell{\textbf{Semantic-Payload } \\ \textbf{Steganography (Ours)}} \\
    \midrule
    \textbf{Payload} & Bitstream & Natural-language message \\
    \textbf{Role of \textit{semantics}} & Carrier property & Payload content \\
    \textbf{Cover media} & Natural-language text & Images/other cover media \\
    \textbf{Specific task} & Linguistic steganography & Sentence-to-image steganography \\
    \bottomrule
  \end{tabular}
\end{table}

\subsection{Problem Definition}
We define semantic-payload steganography as the task of hiding natural-language messages in cover media with three objectives: message recovery, limited perceptual distortion, and low detectability under steganalysis.

Formally, let $m$ denote a natural language message, and $O_{\text{cover}}$ denote a cover object. The encoder $E(\cdot, \cdot)$ embeds $m$ into $O_{\text{cover}}$, producing the stego object $O_{\text{stego}} = E(O_{\text{cover}}, m)$. The corresponding decoder $D(\cdot)$ aims to recover the original message as $\hat{m} = D(O_{\text{stego}})$. The semantic-payload steganography task must satisfy the following constraints:
\begin{equation*}
  \begin{cases}
  \textbf{Imperceptibility:} & \Delta(O_{\text{cover}}, O_{\text{stego}}) \le \epsilon, \\
  \textbf{Undetectability:} & SA(O_{\text{stego}}) \approx SA(O_{\text{cover}}),\\
  \textbf{Recoverability:} & Sim(\hat{m}, m) \rightarrow 1.
  \end{cases}
  \label{eq:semantic_constraints}
\end{equation*}  
Specifically, recoverability measures the semantic consistency of the recovered message using the similarity function $Sim(\cdot, \cdot)$. Imperceptibility evaluates the similarity between cover and stego objects, where $\Delta(\cdot, \cdot)$ measures perceptual distortion, and $\epsilon$ denotes the maximum allowable perceptual difference. Steganalysis undetectability reflects the indistinguishability between cover and stego objects against steganalysis, with $SA(\cdot)$ denoting the output of the steganalysis tool. 


\subsection{Benchmark}
There is an absence of datasets specially designed for evaluating semantic-payload steganography. To fill this gap, we construct the \textbf{Invisible Text (IVT)} dataset using diverse natural-language texts drawn from existing datasets as secret messages. We divide the messages into three levels of granularity: short messages (e.g. phrases or sentences), medium messages (e.g. several sentences), and long messages (e.g. paragraphs), denoted as IVT-S, IVT-M, and IVT-L, respectively. We further construct IVT\textsuperscript{G}, whose messages are all newly synthesized by DeepSeek \cite{deepseek2024}. IVT\textsuperscript{G} follows the same three granularity levels as IVT.

\subsection{Sentence-to-Image Steganography}
Specifically, we formalize the sentence-to-image steganography task as a concrete instantiation of the semantic-payload steganography. Following the definition above, the object $O$ in this task is specified as an image $I$. The objective is to secretly embed natural-language messages within the image while ensuring that the hidden information can be recovered.

\section{Method}
\label{sec:method}
\subsection{Representing Information as Tokens}

\begin{figure*}[tb]
    \centering
    \includegraphics[width=0.99\textwidth]{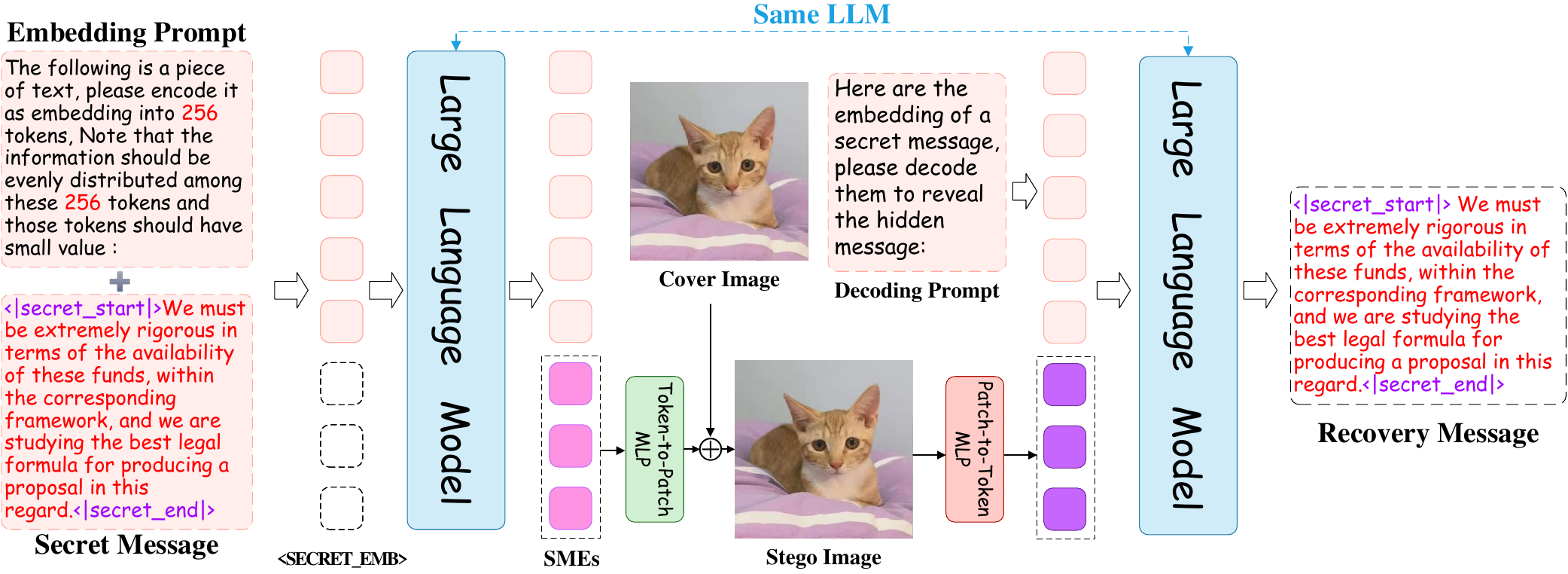}
    \caption{Pipeline of the $\mathrm{S^2LM}$ framework. We use an embedding prompt to guide the LLM to encode the secret message into secret message embeddings (SMEs). The token-to-patch MLP maps the SMEs to a patch representation, which is reshaped into an image embedding and added to the cover image to form the stego image. The patch-to-token MLP converts patches of the stego image into token embeddings. We use a decoding prompt to guide the same LLM to autoregressively recover the secret message from these embeddings.}
    \label{fig:figure2}
\end{figure*}

The representation of secret message is a critical aspect of semantic-payload steganography. Existing methods typically represent secret payloads as binary bitstreams \cite{tancik2020stegastamp,zhu2018hidden,wengrowski2019light} However, such low-level representations have two significant limitations, as discussed in \cite{mielke2021between}:
First, bitstreams do not explicitly preserve the linguistic structure and contextual dependencies of natural-language messages. 
Second, encoding a sentence as a bitstream leads to an excessively long sequence that is difficult to process. Therefore, it is necessary to explore a more effective representation for sentence-level information.

To address these issues, we leverage a tokenizer to convert the secret sentence into tokens. Tokenization \cite{webster1992tokenization} is a widely used technique in NLP, where sentences are broken down into smaller units called \textbf{\textit{Tokens}}. Unlike character-level encoding, tokens correspond to semantically meaningful units, allowing a more compact representation. Compared with bitstreams, token sequences provide a more compact representation and enable the LLM to model contextual dependencies among textual units, improving the efficiency and stability of subsequent embedding and decoding.

\subsection{LLMs for Steganography}
As shown in Fig.~\ref{fig:figure2}, the $\mathrm{S^2LM}$ consists of three key modules: a pre-trained LLM for embedding and decoding the secret information, a Token-to-Patch MLP $\mathcal F_{T2P}$ for aligning text features and image features, and a Patch-to-Token MLP $\mathcal F_{P2T}$ for embedding visual features to token embeddings. Notably, $\mathrm{S^2LM}$ is a general framework that can be instantiated with different pre-trained LLMs.

\noindent
\textbf{Semantic Embedding via LLMs.}
The LLM is recognized as the \textit{de facto} standard architecture in the NLP community. Recent studies, however, have demonstrated that LLM can serve as a universal interface with different instructions in language to complete different tasks. In the computer vision community, researchers have explored integrating LLMs with vision backbones for segmentation \cite{Lai2023LISARS}, grounding \cite{xu2024pixel,peng2024grounding,kang2025your}, and other visual tasks \cite{zeng2024X}. Inspired by these works, we employ an LLM to generate embeddings conditioned by secret information, which will be inserted into the cover image to produce the stego image. Owing to the strong generalization ability of LLMs, we are able to compress the semantic information into a few embeddings.

Specifically, as illustrated in Fig.~\ref{fig:figure2}, we design an embedding prompt $X_{prompt}^{embed}$ and concatenate it with a secret message to instruct the LLM to embed the secret message into embeddings, which we call secret message embeddings (SMEs). We call this strategy \textbf{Prompt-Guided Embedding}. The SMEs are then aligned with the image feature using the $\mathcal F_{T2P}$. After that, we reshape it and add to the cover image to get a stego image. 

\noindent
\textbf{Universal Insertion.} To insert SMEs into the cover image, we use $\mathcal F_{T2P}$ to align the token embeddings with the image features and directly add the resulting embedding to the cover image to produce a stego image \cite{zhang2020udh}. In contrast to prior approaches generate the stego image conditioned on both the cover image and the secret message, this method decouples the secret message from the carrier, which empirically results in smoother optimization and better performance.

\noindent
\textbf{Autoregressive Decoding via LLMs.}
After receiving the stego image, we reshape it in patches and use $\mathcal F_{P2T}$ to extract features and concatenate it with decoding prompt $X_{prompt}^{decode}$, which is used to instruct the LLM to decode the secret message. Unlike existing  methods that independently decode each secret bit, the LLM recovers the secret message autoregressively. At each decoding step, self-attention jointly leverages the stego features and previously decoded tokens to predict the next secret token, allowing the recovered context to guide subsequent decoding. Finally, we get the recovery message from the LLM.

\begin{figure*}[t] \centering
    \includegraphics[width=1\textwidth]{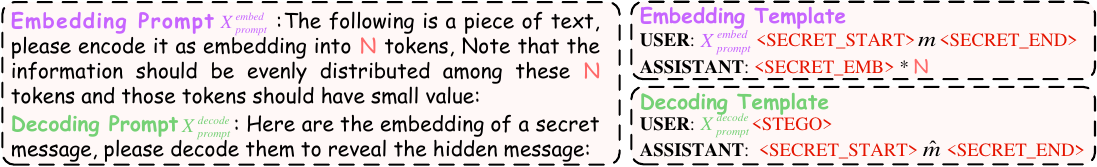}
    \caption{Prompt templates used in the embedding and decoding procedures of $\mathrm{S^2LM}$.} 
    \label{fig:template}
\end{figure*}

\begin{figure*}[tb] \centering
    \includegraphics[width=1\textwidth]{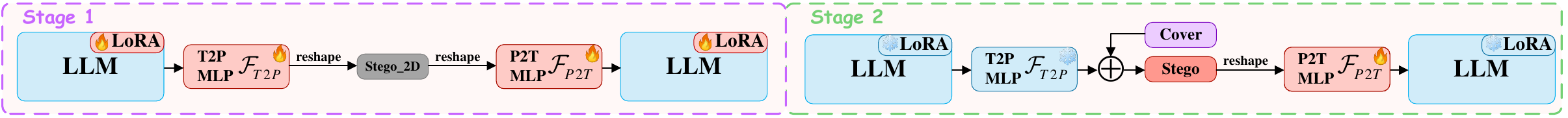}
    \caption{Two-stage training strategy for $\mathrm{S^2LM}$.} 
    \label{fig:training_strategy}
\end{figure*}

\noindent
\textbf{Template Designing.}
As illustrated in Fig.~\ref{fig:template}, we carefully design the embedding and decoding prompts to instruct the LLM to embed and recover the secret message. We introduce two special tokens: $\langle \texttt{SECRET\_START} \rangle$ and $\langle \texttt{SECRET\_END} \rangle$, which are used to separate secret messages from the prompt. We use the output embeddings of the LLM as SMEs. For clarity, we denote the location of the SMEs in the output of LLM as $\langle \texttt{SECRET\_EMB} \rangle$. During the decoding process, the extracted feature will be concatenated with the decode prompt as the input of LLM. We use $\langle \texttt{STEGO} \rangle$ as the placeholder for the feature extracted from stego image by $\mathcal F_{P2T}$. 

\subsection{Training of \texorpdfstring{$\mathit{S^{2}LM}$}{S2LM}}
\label{sec:training}
\noindent
\textbf{Trainable Parameters.}
To preserve the knowledge of the LLM, we use LoRA \cite{hu2022lora} for fine-tuning. Additionally, since we introduce new tokens into the vocabulary (\textit{e.g.}, $\langle \texttt{SECRET\_START} \rangle$), we set the LLM token embeddings ($embed\_tokens$) and the LLM head ($lm\_head$) trainable, together with $\mathcal F_{T2P}$ and $\mathcal F_{P2T}$.

\noindent
\textbf{Stage 1: Pre-Training for Steganography.}
As shown in Fig.~\ref{fig:training_strategy}, the $\mathrm{S^2LM}$ training process consists of two stages. In the first stage, we fine-tune the LLM with LoRA, while training $\mathcal F_{T2P}$ and $\mathcal F_{P2T}$ from scratch. In this stage, we \textbf{DO NOT} introduce the cover image. We wrap the secret message $m$ with $\langle \texttt{SECRET\_START} \rangle$ and $\langle \texttt{SECRET\_END} \rangle$ to construct the input sequence $M$:
\begin{align}
    \label{equation:encode_1}
    M = \langle \texttt{SECRET\_START} \rangle \,\|\, m \,\|\, \langle \texttt{SECRET\_END} \rangle.
\end{align}
Given the embedding prompt $X_{prompt}^{embed}$ along with $M$, we feed them into the LLM, which outputs a response token sequence $Y_{res}$, which can be formulated as:
\begin{align}
    \label{equation:encode_2}
    Y_{res} = LLM(X_{prompt}^{embed}, M).
\end{align}
We extract the last-layer embedding $e \in \mathbb{R}^{N\times d_{emb}}$ of the LLM as the secret embedding corresponding to the $\rm \langle SECRET\_EMB \rangle$ token in $Y_{res}$ and apply $\mathcal F_{T2P}$ to obtain $p \in \mathbb{R}^{N\times d_{patch}}$, which aligns with the image patch dimension. We note that $d_{patch}=P^2\cdot C$, $(P,P)$ is the resolution of each image patch, and $N=HW/P^2$ \cite{dosovitskiy2020image}. Subsequently, $p$ will be processed by $\mathcal F_{P2T}$ to align with the token embedding dimension of the LLM, yielding $e' \in \mathbb{R}^{N\times d_{emb}}$. To prevent the LLM from focusing solely on a subset of embeddings, we randomly mask $e'$ with a mask ratio $R$, which we refer to as the \textbf{Mask Strategy}, by modifying the attention mask of the LLM input. Finally, we concatenate it with decode prompt $X_{prompt}^{decode}$ and feed it into the LLM to obtain the recovery message $\hat{M}$:
\begin{equation}
    \begin{gathered}
        p  = \mathcal F_{T2P}(e), \quad 
        e' = \mathcal F_{P2T}(p), \\
        \hat{M} = LLM(X_{prompt}^{decode}, \mathrm{Mask}(e', R)).
    \end{gathered}
\end{equation}    
\noindent
\textbf{Stage 2: Fine-Tuning on Stego Images.}
In the second stage, we \textbf{FREEZE} the LLM and $\mathcal F_{T2P}$, \textbf{TRAIN} $\mathcal F_{P2T}$, and introduce the cover image $I_{cover} \in \mathbb{R}^{C\times H\times W}$. After obtaining $p \in \mathbb{R}^{N\times d_{patch}}$, we reshape it from patch representation to the image representation $I_{emb} \in \mathbb{R}^{C\times H\times W}$, and add it to $I_{cover}$ to obtain stego image $I_{stego}$. The entire process can be formulated as follows:
\begin{equation}
    \label{equation:embedding}
    \begin{aligned}
        I_{emb} &= Reshape(p), \\
        I_{stego} &= I_{cover} + I_{emb}.
    \end{aligned}
\end{equation}    
After receiving the stego image, we divide the it into patches to transform it into a 1D sequence $\hat p\in \mathbb{R}^{N\times d_{patch}}$ and feed it into $\mathcal F_{P2T}$ to obtain $\hat e'\in \mathbb{R}^{N\times d_{emb}}$. Finally, we concatenate decode prompt $X_{prompt}^{decode}$ with the $\hat{e}'$ and feed it into the LLM to get the recovery message $\hat{M}$, which can be formulated as follows:
\begin{equation}
    \label{equation:phase2}
    \begin{gathered}
        \hat p = Patchify(I_{stego}), \quad 
        \hat e' = \mathcal F_{P2T}(\hat p), \\
        \hat{M} = LLM(X_{prompt}^{decode}, \hat e').
    \end{gathered}
\end{equation}    
\noindent
\textbf{Training Objectives.} 
The $\mathrm{S^2LM}$ is trained using the text generation loss $\mathcal L_{txt}$ and the embedding loss $\mathcal L_{emb}$. The overall objective $\mathcal L$ of each stage is defined as the weighted sum of these losses:
\begin{equation}
    \label{equation:total_loss}
    \begin{aligned}
        \mathcal{L_\text{stage-1}} &= \lambda_{txt} \mathcal{L}_{txt} + \lambda_{emb} \mathcal{L}_{emb}, \\
        \mathcal{L_\text{stage-2}} &= \lambda_{txt} \mathcal{L}_{txt}.
    \end{aligned}
\end{equation}
Specifically, $\mathcal L_{txt}$ is the cross-entropy (CE) loss for text generation, and $\mathcal L_{emb}$ is the L1 loss, which encourages the model to introduce minimal artifacts in the cover image. Those losses can be formulated as:
\begin{eqnarray}
    \label{equation:loss_function}
    \mathcal{L}_{txt} = \textbf{CE}(\hat{M}, M), \ \mathcal{L}_{emb} = \textbf{L1}(p, 0).
\end{eqnarray}
\section{Experiments}
\label{sec:experiment}

\subsection{Threat Model}
Following the warden model originating from Simmons' Prisoners' Problem and subsequently formalized in treatments of steganographic security \cite{simmons1984prisoners,cachin1998information,anderson1998limits}, we adopt a covert communication scenario involving a \textbf{sender} (Alice), a \textbf{receiver} (Bob), and a \textbf{warden} (Eve). This formulation also underlies modern neural steganography \cite{baluja2017hiding,zhu2018hidden,lin2024zero,tancik2020stegastamp,jing2021hinet}.

\noindent
\textbf{Communication Settings. } Given a cover image $I_{\mathrm{cover}}$ and a secret message $m$, Alice employs a steganographic encoder $E$ to generate a stego image and then transmits $I_{\mathrm{stego}}$ to Bob through a public communication channel. Bob uses the corresponding decoder $D$ to recover the hidden message. Meanwhile, Eve monitors the public channel and observes the transmitted images.

\noindent
\textbf{Security Objective.}
The security objective considered in this work is steganographic undetectability against a passive warden. Eve observes the images transmitted through the public channel and performs binary hypothesis testing between $H_0: I \sim P_{\mathrm{cover}}$ and $H_1: I \sim P_{\mathrm{stego}}$ \cite{cachin1998information,wang2008perfectly}. To operationalize this threat model, we instantiate Eve using statistical and target-specific neural steganalysis tool and evaluate their ability to distinguish stego images from cover images. Their detection performance serves as an empirical measure of resistance to steganalysis.

\noindent
\textbf{Adversary's Knowledge and Capabilities.} Following the conventional setting, Eve monitor all traffic transmitted through the public communication channel and does not modify or corrupt the transmitted images. Passivity restricts Eve's interaction with the communication channel, rather than her knowledge or detection capability. Following principle for data-hiding security \cite{cayre2008kerckhoffs}, we assume that Eve may also obtain a limited number of labeled cover and stego samples generated by the target steganographic method and use them to train a target-specific steganalysis model. The model architecture and training procedure are public, whereas the parameters of the specific trained encoder/decoder instance remain secret and are inaccessible to Eve.

\subsection{Experimental Settings}
\label{sec:experimentalsetting}

\noindent
\textbf{Model Variants.}
$\mathrm{S^2LM}$ is a general framework compatible with different pre-trained LLM. In this paper, we instantiate it with four models: Llama3.2-1B \cite{grattafiori2024llama}, Qwen2.5-0.5B \cite{qwen2025qwen25technicalreport}, Gemma3-1B \cite{team2025gemma}, and MiniCPM-1B \cite{hu2024minicpm}. We chose edge-deployable language models to guarantee lightweight operation while maintaining the scalability to larger models.

\noindent
\textbf{Implementation Details.}
We employ AdamW optimizer \cite{loshchilov2017decoupled}. The weights $\lambda_{txt}$ and $\lambda_{emb}$ are set to 1.0 and 1.0, respectively. The mask ratio $R$ is set to 0.25. We set the patch size $P=16$, resulting in $N=256$ for a $256 \times 256$ image. We use Universal Suppressor \cite{Wu2025TheDI} for regularizing training to simulate practical image transmission, including 8-bit quantization and pixel-value clipping.

\noindent
\textbf{Datasets.}
For the secret message, we use Wanjuan 1.0 \cite{He2023WanJuanAC}, a comprehensive corpus designed for large language model pre-training. We randomly select 320,000 samples with fewer than 1,000 words as secret messages for training. For the cover image, we use the COCO training dataset \cite{lin2014microsoft}. We evaluate the performance of steganographic models on the sentence-to-image steganography task, using the IVT dataset as secret messages and the COCO test dataset as cover images. All images are centrally cropped to $256 \times 256$ during both training and evaluation. We cannot determine whether the LLMs were exposed to IVT samples at any stage of model development, including pre-training, fine-tuning, or reinforcement learning from human feedback (RLHF). However, we ensure that IVT does \textbf{not overlap} with the data used to train $\mathrm{S^2LM}$. We additionally evaluate on IVT\textsuperscript{G}, which consists entirely of newly generated messages. For more details about the dataset, please refer to the Appendix.

\noindent
\textbf{Evaluation Metrics.}
The performance of the steganographic model is evaluated from three complementary perspectives: \textbf{recoverability}, \textbf{imperceptibility}, and \textbf{steganalysis undetectability}.
For recoverability, we evaluate the recovered text using Word Error Rate (WER), BLEU-4 \cite{papineni2002bleu}, ROUGE-L \cite{lin2004rouge}, BERT-Score \cite{zhang2019bertscore}, and Exact Recovery Rate (ERR), following established evaluation practice for assessing the lexical and semantic fidelity of recovered text \cite{zhao2025sok}. Specifically, WER, BLEU-4, and ROUGE-L are lexical-level metrics that rely on exact word or word-sequence matching between the recovered and original texts. Thus, only correctly recovered lexical content contributes to these metrics, while semantically similar but lexically different words are still regarded as mismatches. ERR provides a stricter sample-level evaluation by measuring the proportion of secret messages that are recovered completely without any error. A sample is counted as correct only when the entire recovered message exactly matches the original message. In contrast, BERT-Score evaluates similarity at the semantic level based on contextual representations. The combination of these metrics therefore evaluates recoverability from complementary perspectives, covering lexical accuracy, exact recovery, and semantic consistency.
For imperceptibility, we adopt PSNR and SSIM to measure the visual similarity between the cover and stego images.
For empirical steganalysis evaluation, we employ SOTA steganalysis tools including StegExpose \cite{Boehm2014StegExposeA}, SRNet \cite{Boroumand2019DeepRN}, PENet \cite{wei2024color}, and DBS2Net \cite{hu2024lightweight} to quantify the empirical detectability of stego images.

\noindent
\textbf{Baselines.}
We compare $\mathrm{S^2LM}$ with different methods, including  StegaStamp \cite{tancik2020stegastamp}, LanNet \cite{lan2023robust}, CRMark \cite{chen2025learning}, and FPGP \cite{zhang2025fpgp}.

\subsection{Sentence-to-Image Steganography Results}
All methods are evaluated using the same IVT secret messages, the same cover images at a fixed resolution, and identical data splits. For each message-cover pair, each method is assessed end to end by its ability to embed the message into the cover image and recover it from the resulting stego image. The internal representation of the message, whether a bitstream, a token sequence, or a learned continuous representation, is a design choice specific to each method and is not the focus of our comparison.

\noindent
\textbf{Results of Existing Methods.}
Existing methods are not suitable for the sentence-to-image steganography task, which fundamentally differs in that it requires encoding a substantial amount of semantic information. As shown in Tab.~\ref{table:main_result}, StegaStamp fails to recover secret messages in IVT-S, while FPGP perform well only on IVT-S, but fail to recover messages in IVT-M, not to mention handling IVT-L. However, LanNet and CRMark also fail on IVT-L. 

We retrain all of the baselines for each subset of IVT dataset. However, we observe a severe mode collapse problem, which is consistent with the phenomenon reported in \cite{petrov2025we}. We provide a detailed analysis in Appendix. This issue primarily stems from limitations in the architectural design and bitstream representation of secret information. As a result, these models are restricted to hiding bit-level information and cannot handle semantic-level content.

\noindent
\textbf{Results of $\mathbf{S^2LM}$.}
Existing methods fail to address this task effectively, whereas $\mathrm{S^2LM}$ successfully handles secret messages containing complex and semantically rich natural language information. 

\begin{table*}[]
    \centering
    \scriptsize
    \tabcolsep=2pt
    \caption{Quantitative results on the IVT-S, IVT-M, and IVT-L benchmarks.}
    \begin{tabular}{l|cccccc|cccccc|cccccc}
        \toprule
        \multicolumn{1}{c|}{\multirow{3}{*}{\textbf{Methods}}}                  & \multicolumn{6}{c|}{\textbf{IVT-S}}                                                                                                 & \multicolumn{6}{c|}{\textbf{IVT-M}}                                                                                                 & \multicolumn{6}{c}{\textbf{IVT-L}}                                                                                                 \\ \cmidrule{2-19} 
        \multicolumn{1}{c|}{}                                          & \multicolumn{4}{c|}{\textbf{Secret/Recovery}}                                           & \multicolumn{2}{c|}{\textbf{Cover/Stego}} & \multicolumn{4}{c|}{\textbf{Secret/Recovery}}                                           & \multicolumn{2}{c|}{\textbf{Cover/Stego}} & \multicolumn{4}{c|}{\textbf{Secret/Recovery}}                                           & \multicolumn{2}{c}{\textbf{Cover/Stego}} \\ \cmidrule{2-19} 
        \multicolumn{1}{c|}{}                                          & \textbf{WER}   & \textbf{BLEU}  & \textbf{ROUGE} & \multicolumn{1}{c|}{\textbf{BERT-S}} & \textbf{PSNR}       & \textbf{SSIM}       & \textbf{WER}   & \textbf{BLEU}  & \textbf{ROUGE} & \multicolumn{1}{c|}{\textbf{BERT-S}} & \textbf{PSNR}       & \textbf{SSIM}       & \textbf{WER}   & \textbf{BLEU}  & \textbf{ROUGE} & \multicolumn{1}{c|}{\textbf{BERT-S}} & \textbf{PSNR}      & \textbf{SSIM}       \\ \midrule
        \textbf{StegaStamp (CVPR 20)}                                            & 0.994          & 0.002          & 0.002          & \multicolumn{1}{c|}{0.120}           & 40.4                & 0.927               & -              & -              & -              & \multicolumn{1}{c|}{-}               & -                   & -                   & -              & -              & -              & \multicolumn{1}{c|}{-}               & -                  & -                   \\
        \textbf{FPGP (TCSVT 26)}                                                  & 0.150          & 0.882          & 0.925          & \multicolumn{1}{c|}{0.914}           & 43.5                & 0.954               & 0.958          & 0.096          & 0.069          & \multicolumn{1}{c|}{0.124}           & 36.4                & 0.885               & -              & -              & -              & \multicolumn{1}{c|}{-}               & -                  & -                   \\
        \textbf{LanNet (AAAI 23)}                                                & {\underline{0.004}}    & \textbf{0.906}    & \textbf{0.997} & \multicolumn{1}{c|}{\textbf{0.996}}  & 42.1                & 0.951               & 0.244          & 0.698          & 0.836          & \multicolumn{1}{c|}{0.787}           & 41.3                & 0.949               & 0.995          & 0.031          & 0.088          & \multicolumn{1}{c|}{0.130}           & 40.9               & 0.941               \\
        \textbf{CRMark (ICCV 25)}                                                & \textbf{0.003} & {\underline{0.904}} & \underline{0.994}          & \multicolumn{1}{c|}{{\underline{0.994}}}     & \textbf{47.3}       & {\underline{0.985}}         & 0.185          & 0.724          & 0.862          & \multicolumn{1}{c|}{0.894}           & 40.5                & 0.914               & 0.972          & 0.042          & 0.105          & \multicolumn{1}{c|}{0.169}           & 37.6               & 0.825               \\
        $\text{\textbf{S\textsuperscript{2}LM}}$\textbf{-Qwen2.5-0.5B} & 0.045          & 0.874          & 0.887          & \multicolumn{1}{c|}{0.919}           & 40.1                & 0.956               & \textbf{0.043} & \textbf{0.939} & \textbf{0.974} & \multicolumn{1}{c|}{0.952}           & 38.6                & {\underline{0.973}}         & \textbf{0.083} & \textbf{0.903} & \textbf{0.944} & \multicolumn{1}{c|}{\textbf{0.945}}  & 38.3               & 0.927               \\
        $\text{\textbf{S\textsuperscript{2}LM}}$\textbf{-MiniCPM-1B}   & 0.046          & 0.844          & 0.923          & \multicolumn{1}{c|}{0.959}           & 41.6                & 0.970               & {\underline{0.091}}    & 0.889          & {\underline{0.956}}    & \multicolumn{1}{c|}{{\underline{0.953}}}     & {\underline{41.4}}          & 0.969               & 0.172          & 0.801          & 0.874          & \multicolumn{1}{c|}{0.900}           & {\underline{41.0}}         & \textbf{0.971}      \\
        $\text{\textbf{S\textsuperscript{2}LM}}$\textbf{-Llama3.2-1B}  & 0.142          & 0.837          & 0.892          & \multicolumn{1}{c|}{0.939}           & {\underline{44.9}}          & \textbf{0.987}      & 0.095          & {\underline{0.918}}    & 0.950          & \multicolumn{1}{c|}{0.952}           & \textbf{45.0}       & \textbf{0.987}      & {\underline{0.109}}    & {\underline{0.883}}    & {\underline{0.922}}    & \multicolumn{1}{c|}{{\underline{0.933}}}     & \textbf{41.9}      & {\underline{0.962}}         \\
        $\text{\textbf{S\textsuperscript{2}LM}}$\textbf{-Gemma3-1B}    & 0.100          & 0.801          & 0.746          & \multicolumn{1}{c|}{0.834}           & 42.4                & 0.973               & {\underline{0.091}}    & 0.895          & 0.942          & \multicolumn{1}{c|}{\textbf{0.956}}  & 40.8          & 0.965              & 0.211          & 0.807          & 0.880          & \multicolumn{1}{c|}{0.887}           & 36.5               & 0.915               \\ \bottomrule   
    \end{tabular}
    \label{table:main_result}
\end{table*}

\begin{figure*}[tb] \centering
    \includegraphics[width=1\textwidth]{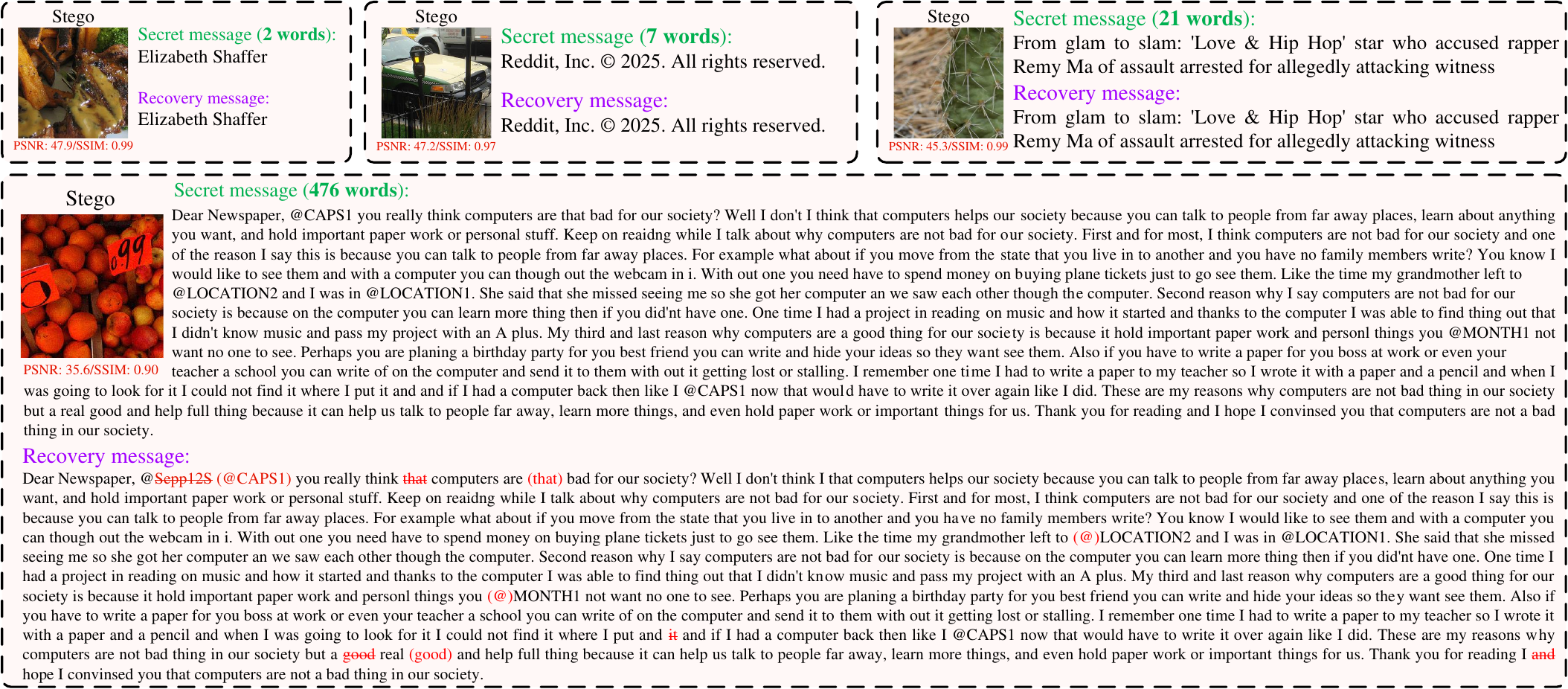}
    \caption{Qualitative results of $\mathrm{S^2LM}$-Qwen2.5-0.5B on different lengths of secret messages.} 
    \label{fig:quantitative_result}
\end{figure*}

\begin{table}[t]
    \centering
    \begingroup
    \color{black}
    \small
    \tabcolsep=7pt
    \caption{Exact recovery rate on IVT-S, IVT-M, and IVT-L.}
    \begin{tabular}{l|ccc}
        \toprule
        \textbf{Methods} & \textbf{IVT-S} & \textbf{IVT-M} & \textbf{IVT-L} \\ \midrule
        $\mathrm{S^2LM}$-Qwen2.5-0.5B & 89.5\% & 83.6\% & 81.8\% \\
        StegaStamp & 0.0\% & - & - \\
        FPGP & 76.2\% & 0.0\% & - \\
        LanNet & 92.2\% & 36.8\% & 0.0\% \\
        CRMark & 93.7\% & 32.3\% & 0.0\% \\ \bottomrule
    \end{tabular}
    \label{tab:exact_recovery_rate}
    \endgroup
\end{table}

On the IVT-S, all four $\mathrm{S^2LM}$ variants demonstrate competitive performance compared to traditional baselines in secret recovery while maintaining high visual quality. Specifically, $\mathrm{S^2LM}$-MiniCPM-1B achieves the highest ROUGE-L and BERT-Score among the four variants, while other variants also deliver competitive results. On the IVT-M, which involves more complex messages, our $\mathrm{S^2LM}$ variants significantly outperform prior methods. $\mathrm{S^2LM}$-Qwen2.5-0.5B achieves the best recovery performance among all variants while maintaining high visual quality. $\mathrm{S^2LM}$-MiniCPM-1B and $\mathrm{S^2LM}$-Gemma3-1B exhibit comparable performance. On the IVT-L, all models exhibit performance degradation. $\mathrm{S^2LM}$-Qwen2.5-0.5B get the top recovery performance, $\mathrm{S^2LM}$-MiniCPM-1B and $\mathrm{S^2LM}$-Llama3.2-1B demonstrate moderate performance declines. Moreover, the ERR results in Tab.~\ref{tab:exact_recovery_rate} show that $\mathrm{S^2LM}$-Qwen2.5-0.5B can recover entire secret messages without error, although exact recovery remains more challenging on IVT-M and IVT-L.

Experimental results demonstrate that all $\mathrm{S^2LM}$ variants achieve strong performance on both IVT-S and IVT-M. However, for these models, IVT-L remains challenging, which comprises multi-sentence structures characterized by complex syntactic patterns and diverse lexical content. Additional experiments in the Appendix show that $\mathrm{S^2LM}$-Qwen2.5-0.5B, trained with perturbation augmentation, maintains strong message recovery performance under six image distortions. However, the results demonstrate that $\mathrm{S^2LM}$ is a flexible framework applicable to different LLMs.

Although the secret messages in IVT were not used to train $\mathrm{S^2LM}$, they were drawn from public datasets and may have appeared in the backbone LLMs' pre-training corpora. We therefore constructed IVT\textsuperscript{G} using texts generated specifically for this evaluation through the DeepSeek API \cite{deepseek2024}. Following IVT, we divided the messages into short, medium, and long subsets. As shown in Table \ref{table:IVT_G}, without further training, $\mathrm{S^2LM}$-Qwen2.5-0.5B achieved WERs of \textbf{0.040}, \textbf{0.034}, and \textbf{0.047} on the three subsets, respectively, while maintaining PSNR above \textbf{39} dB.



\noindent
\textbf{Qualitative Results.} As depicted in Fig.~\ref{fig:quantitative_result}, we provide a visualization of stego images and recovery results of $\mathrm{S^2LM}$-Qwen2.5-0.5B in different lengths of information. We mark the difference in red. As we can see, the recovery message and the ground truth keep a high semantic similarity, while the stego image still has a good visual quality. More samples can be found in Appendix.

\begin{table*}[t]
    \centering
    \caption{Quantitative results on the IVT\textsuperscript{G}-S, IVT\textsuperscript{G}-M, and IVT\textsuperscript{G}-L benchmarks.}
    \scriptsize
    \tabcolsep=2.2pt
    \begin{tabular}{@{}l|cccccc|cccccc|cccccc@{}}
        \toprule
        \multicolumn{1}{c|}{\textbf{}}                                 & \multicolumn{6}{c|}{\textbf{IVT\textsuperscript{G}-S}}                                                                                              & \multicolumn{6}{c|}{\textbf{IVT\textsuperscript{G}-M}}                                                                                              & \multicolumn{6}{c}{\textbf{IVT\textsuperscript{G}-L}}                                                                                              \\ \cmidrule(l){2-19} 
        \multicolumn{1}{c|}{\textbf{Methods}}                          & \multicolumn{4}{c|}{\textbf{Secret/Recovery}}                                        & \multicolumn{2}{c|}{\textbf{Cover/Stego}} & \multicolumn{4}{c|}{\textbf{Secret/Recovery}}                                        & \multicolumn{2}{c|}{\textbf{Cover/Stego}} & \multicolumn{4}{c|}{\textbf{Secret/Recovery}}                                        & \multicolumn{2}{c}{\textbf{Cover/Stego}} \\ \cmidrule(l){2-19} 
        \multicolumn{1}{c|}{\textbf{}}                                 & \textbf{WER} & \textbf{BLEU} & \textbf{ROUGE} & \multicolumn{1}{c|}{\textbf{BERT-S}} & \textbf{PSNR}       & \textbf{SSIM}       & \textbf{WER} & \textbf{BLEU} & \textbf{ROUGE} & \multicolumn{1}{c|}{\textbf{BERT-S}} & \textbf{PSNR}       & \textbf{SSIM}       & \textbf{WER} & \textbf{BLEU} & \textbf{ROUGE} & \multicolumn{1}{c|}{\textbf{BERT-S}} & \textbf{PSNR}       & \textbf{SSIM}      \\ \midrule
        \multicolumn{1}{l|}{$\text{\textbf{S\textsuperscript{2}LM}}$\textbf{-Qwen2.5-0.5B}} & {\underline{0.040}}    & 0.911          & 0.944          & \multicolumn{1}{c|}{0.953}          & 41.6           & 0.960          & \textbf{0.034}          & {\underline{0.933}}    & {\underline{0.963}}    & \multicolumn{1}{c|}{{\underline{0.966}}}    & 39.1          & \multicolumn{1}{c|}{0.984}                & {\underline{0.047}}    & {\underline{0.916}}    & {\underline{0.959}}    & \multicolumn{1}{c|}{{\underline{0.958}}}    & 40.2          & 0.968          \\
        \multicolumn{1}{l|}{$\text{\textbf{S\textsuperscript{2}LM}}$\textbf{-MiniCPM-1B}}   & {\underline{0.040}}    & \textbf{0.935} & \textbf{0.972} & \multicolumn{1}{c|}{\textbf{0.977}} & 42.4           & {\underline{0.988}}    & 0.040          & \textbf{0.939} & \textbf{0.973} & \multicolumn{1}{c|}{\textbf{0.976}} & 42.5          & \multicolumn{1}{c|}{{\underline{0.987}}}          & 0.054          & \textbf{0.919} & \textbf{0.963} & \multicolumn{1}{c|}{\textbf{0.963}} & {\underline{42.8}}    & {\underline{0.984}} \\
        \multicolumn{1}{l|}{$\text{\textbf{S\textsuperscript{2}LM}}$\textbf{-Llama3.2-1B}}  & \textbf{0.037} & {\underline{0.916}}    & {\underline{0.947}}    & \multicolumn{1}{c|}{{\underline{0.961}}}    & \textbf{45.6}  & \textbf{0.991} & \underline{0.039} & 0.887          & 0.920          & \multicolumn{1}{c|}{0.938}          & \textbf{46.6} & \multicolumn{1}{c|}{\textbf{0.989}}       & \textbf{0.040} & 0.903          & 0.942          & \multicolumn{1}{c|}{0.914}          & \textbf{43.1} & \textbf{0.988}          \\
        \multicolumn{1}{l|}{$\text{\textbf{S\textsuperscript{2}LM}}$\textbf{-Gemma3-1B}}    & 0.067          & 0.767          & 0.819          & \multicolumn{1}{c|}{0.866}          & {\underline{43.5}}     & 0.980          & 0.096          & 0.832          & 0.857          & \multicolumn{1}{c|}{0.893}          & {\underline{43.6}}    & \multicolumn{1}{c|}{0.977}                & 0.157          & 0.852          & 0.864          & \multicolumn{1}{c|}{0.873}          & 39.2          & 0.932    \\ \bottomrule                
    \end{tabular}
    \label{table:IVT_G}
\end{table*}

\begin{figure*}[t] \centering
    \includegraphics[width=0.95\textwidth,]{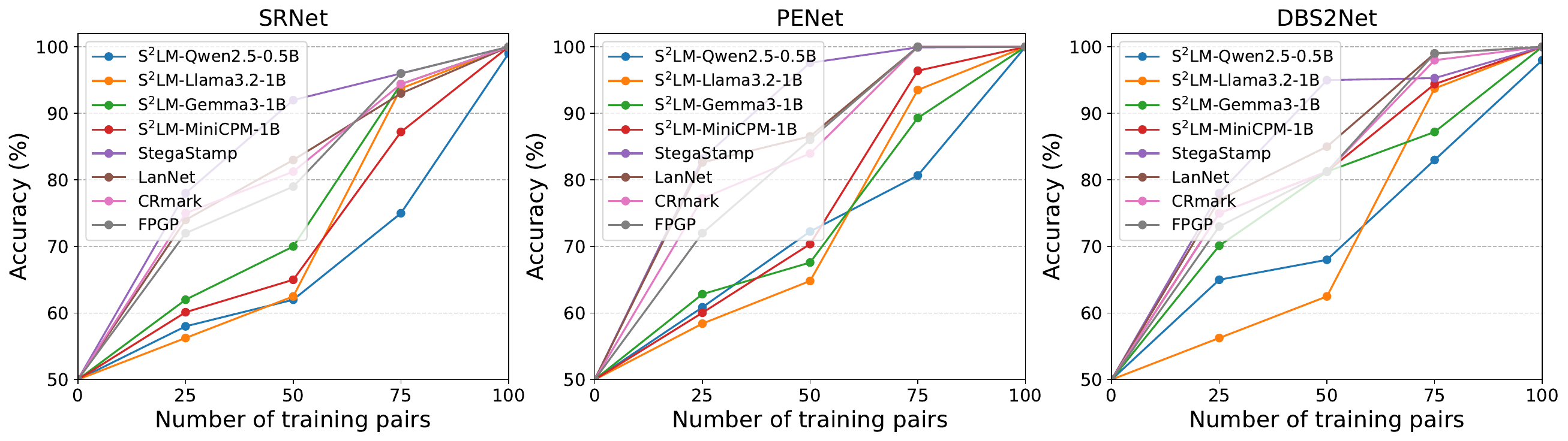}
    \caption{The results of the deep-steganalysis. Note that the closer the accuracy is to 50\%, the stronger its resistance to steganalysis.} 
    \label{fig:deep-analysis}
\end{figure*}

\begin{table}[t]
    \centering
    \caption{Result on the IVT-L of 8-bit quantized $\mathrm{S^2LM}$-Qwen2.5-0.5B.}
    \scriptsize
    \tabcolsep=3pt
    \begin{tabular}{@{}l|cccc|cc@{}}
    \toprule
    \textbf{Methods}          & \textbf{WER} & \textbf{BLEU-4} & \textbf{ROUGE} & \textbf{BERT-S} & \textbf{PSNR} & \textbf{SSIM} \\ \midrule
    $\mathrm{S^2LM}$-Qwen2.5-0.5B-8bit & 0.201        & 0.804           & 0.882          & 0.907           & 40.05         & 0.946         \\ \bottomrule
    \end{tabular}
    \label{tab:Quantization}
\end{table}

\begin{figure}[t]
    \centering
    \includegraphics[width=0.48\textwidth,]{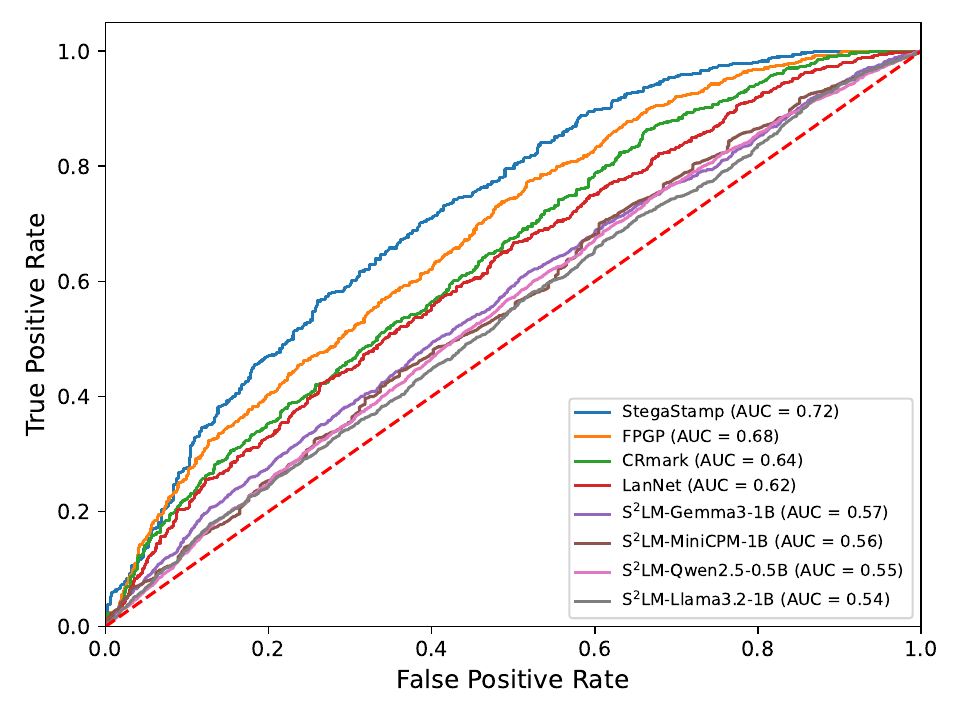}
    \caption{The ROC curve produced by StegExpose.}
    \label{fig:roc}
\end{figure}

\subsection{Model Quantization}
We use a lightweight language model ($\leq 1B$), which runs efficiently on edge devices. After applying 8-bit quantization, its memory usage of the $\mathrm{S^2LM}$-Qwen2.5-0.5B is below 600 MB, making the entire framework both practical and scalable. As shown in Table~\ref{tab:Quantization}, the 8-bit quantized version of $\mathrm{S^2LM}$-Qwen2.5-0.5B processes the IVT-L dataset effectively without any quantization-aware training, with GPU memory usage of less than 2 GB during inference.

\subsection{Steganalysis}
We evaluate anti-steganalysis ability for all the baselines following the standard protocol widely adopted in image steganography \cite{Guan2022DeepMIHDI,jing2021hinet,ke2024stegformer,yu2023cross}.

\noindent
\textbf{Statistical Steganalysis.}
StegExpose \cite{Boehm2014StegExposeA} is a classical steganalysis tool, which integrates several widely used statistical steganalysis methods, including RS analysis \cite{fridrich2001reliable}, Sample Pair analysis \cite{dumitrescu2002detection}, Chi-square attack \cite{westfeld1999attacks}, and Primary Sets analysis \cite{dumitrescu2002steganalysis}, and further improves detection performance through detector fusion. We use StegExpose to evaluate the detectability of stego images produced by $\mathrm{S^2LM}$ and all the  baselines. As shown in Fig.~\ref{fig:roc}, StegExpose yields AUCs of 0.54--0.57 for the four $\mathrm{S^2LM}$ variants, compared with 0.62--0.72 for the evaluated baselines. These results suggest that stego images generated by $\mathrm{S^2LM}$ are harder for StegExpose to detect than those generated by other baselines.

\noindent
\textbf{Deep-Steganalysis.}
In this experiment, Eve is assumed to have access to a limited number of labeled cover/stego pairs in addition to monitoring the public channel. These pairs allow her to train target-specific steganalysis tools, making this setting a stress test under favorable conditions for the attacker. However, obtaining such labeled data can be challenging in practical communication \cite{megias2022subsequent,ker2014steganographer,wang2023linguistic}. Following existing works \cite{Guan2022DeepMIHDI,jing2021hinet,ke2024stegformer,zhu2018hidden,zhang2020udh}, we conduct target-specific steganalysis experiments using cover/stego image pairs generated in the main experiments. For each method, we train SRNet, PENet, and DBS2Net on labeled cover/stego pairs generated by that method. Each detector is evaluated on 1,000 additional cover/stego pairs that do not overlap with its training set. This setting assumes a strong target-specific attacker who knows which steganographic method produces the observed stego images and can obtain fully labeled cover/stego samples from the target method for detector training. We gradually increase the number of leaked training pairs to investigate the detectability of different methods under different amounts of available data, continuing until the steganalysis models can distinguish stego images from cover images.

\begin{table*}[]
    \centering
    \small
    \tabcolsep=6.6pt
    \caption{Ablation study of the module in the $\mathrm{S^2LM}$ framework.}
    \begin{tabular}{c|cccc|cccc|cc}
        \toprule
        \multirow{2}{*}{\textbf{No.}} & \multirow{2}{*}{\textbf{\begin{tabular}[c]{@{}c@{}}Two-Stage\\ Training\end{tabular}}} & \multirow{2}{*}{\textbf{Prompt}} & \multirow{2}{*}{\textbf{\begin{tabular}[c]{@{}c@{}}Embedding\\ Loss\end{tabular}}} & \multirow{2}{*}{\textbf{\begin{tabular}[c]{@{}c@{}}Mask\\ Strategy\end{tabular}}} & \multicolumn{4}{c|}{\textbf{Secret / Recovery}}                        & \multicolumn{2}{c}{\textbf{Cover / Stego}} \\ \cmidrule{6-11} 
        &                                                                                        &                                  &                                                                                    &                                                                                   & \textbf{WER}    & \textbf{BLEU-4} & \textbf{ROUGE-L} & \textbf{BERT-S} & \textbf{PSNR}       & \textbf{SSIM}        \\ \midrule
        1                             &                                                                                        &                                  &                                                                                    &                                                                                   & 0.3162          & 0.5925          & 0.6781           & 0.6816          & 34.13               & 0.8156               \\
        2                             & \checkmark                                                                             &                                  &                                                                                    &                                                                                   & 0.2210          & 0.6819          & 0.7403           & 0.8051          & 32.93               & 0.8460               \\
        3                             & \checkmark                                                                             & \checkmark                       &                                                                                    &                                                                                   & 0.2200          & {\underline{0.7319}}    & {\underline{0.7965}}     & {\underline{0.8295}}    & 31.41               & 0.8248               \\
        4                             & \checkmark                                                                             & \checkmark                       & \checkmark                                                                         &                                                                                   & {\underline{0.2139}}    & 0.5676          & 0.6606           & 0.7465          & \textbf{45.25}      & \textbf{0.9839}      \\
        5                             &                                                                                        & \checkmark                       & \checkmark                                                                         & \checkmark                                                                        & 0.2381          & 0.6703          & 0.7496           & 0.8119          & 34.24               & 0.7820               \\
        6                             & \checkmark                                                                             & \checkmark                       & \checkmark                                                                         & \checkmark                                                                        & \textbf{0.0834} & \textbf{0.9028} & \textbf{0.9435}  & \textbf{0.9453} & {\underline{38.30}}         & {\underline{0.9266}}         \\ \bottomrule       
    \end{tabular}
    \label{tab:ablation}
\end{table*}

As shown in Fig.~\ref{fig:deep-analysis}, the detection accuracy of all methods generally increases as more leaked samples become available. Therefore, we compare different steganographic methods under the same attack model and with the same number of training pairs. Under these controlled settings, $\mathrm{S^2LM}$ variants generally achieve lower detection accuracy than all the  baselines across all three steganalysis models. For example, with 75 training pairs, $\mathrm{S^2LM}$-Qwen2.5-0.5B achieves substantially lower detection accuracy than all evaluated baselines across SRNet, PENet, and DBS2Net. Its detection accuracies are approximately 75\%, 80\%, and 83\%, compared with 93-100\%, 97-100\%, and 95-100\% for the baselines, respectively.

\begin{figure}[t]
    \centering
    \includegraphics[width=0.45\textwidth]{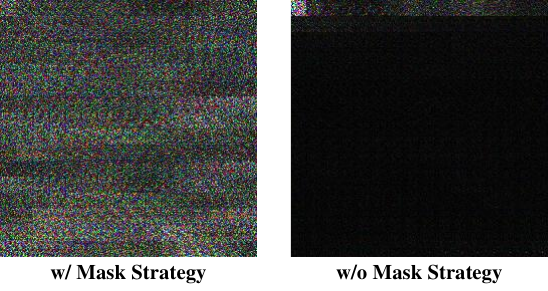}
    \caption{The Visualization of SMEs w/ or w/o mask strategy, which are enhanced by $5\times$ for better visualization.} 
    \label{tab:sem}
\end{figure}

\subsection{Ablation Study}
In this section, we conduct extensive ablation studies on $\mathrm{S^2LM}$-Qwen2.5-0.5B in the IVT-L to reveal the contribution of each component.

\noindent
\textbf{Effectiveness of Two-Stage Training.}
Without the two-stage training strategy, the $\mathrm{S^2LM}$ struggles to recover the secret message (\textbf{No.1} vs. \textbf{No.2}, \textbf{No.5} vs. \textbf{No.6}). That may be because introducing the image modality at an early stage disrupts the LLM to learn steganographic mechanisms.

\noindent
\textbf{Effectiveness of the Prompt.}
In this experiment, we do not use prompts to guide the LLM during the embedding and decoding processes. During embedding, we simply tokenize the secret message and input it into the LLM to obtain SMEs. Similarly, during decoding, we directly feed the features extracted by the Patch-to-Token MLP into the LLM to recover the original message. Comparing experiments \textbf{No.2} and \textbf{No.3} reveals that adding prompts leads to performance improvements of \textbf{7.3\%}, \textbf{7.6\%}, and \textbf{3.0\%} in BLEU-4, ROUGE-L, and BERT-Score, respectively. These results confirm that the prompt effectively guides the LLM in understanding and executing the sentence-to-image task.

\noindent
\textbf{Effectiveness of Embedding Loss.}
A comparison between experiments \textbf{No.3} and \textbf{No.4} reveals the contribution of $\mathcal L_{emb}$. $\mathcal L_{emb}$ is used to encourage the LLM to produce minimal artifacts in the cover image and is applied only in the first stage. The experimental results show that the introduction of $\mathcal L_{emb}$ can improve the embedding quality, with gains of \textbf{13.84 dB} and \textbf{0.159} in the PSNR and SSIM. However, compared with \textbf{No.3}, \textbf{No.4} obtains lower BLEU-4, ROUGE-L, and BERT-Score. This degradation may arise because, without the mask strategy, secret information remains concentrated in a limited subset of SMEs. In this setting, $\mathcal L_{emb}$ further reduces the magnitude of the corresponding embedding residuals, making these concentrated signals more difficult to decode.

\noindent
\textbf{Effectiveness of Mask Strategy.}
In the decoding process of the first stage, we mask some input tokens to encourage the LLM to distribute information across the entire cover image. As shown by the comparison between experiments \textbf{No.4} and \textbf{No.6}, this strategy leads to significant improvements in recovery quality. Specifically, WER decreases by \textbf{0.1305}, while BLEU-4, ROUGE-L, and BERT-Score increase by \textbf{0.3352}, \textbf{0.2829}, and \textbf{0.1988}, respectively. We visualize the SMEs in Fig.~\ref{tab:sem}, which shows that the SMEs are evenly distributed with the mask strategy. 

\section{Conclusion}
In this work, we propose a novel concept of semantic-payload steganography, which aims to hide high-level semantic information in carriers and redefines the traditional paradigm of steganography. Building on this insight, we present sentence-to-image steganography as an instance, which focuses on embedding sentence-level information into images. To achieve this task, we propose $\mathrm{S^2LM}$, a novel steganographic framework that leverages LLMs to embed sentence-level information into images, achieving a feat that previously has been out of reach. Additionally, we present evaluation benchmarks, Invisible Text (IVT) and IVT\textsuperscript{G}, comprising a diverse set of sentence-level texts drawn from existing datasets and generated by LLMs, respectively. We hope our work sheds new light on the advancement of semantic-payload steganography for the entire community.

\bibliographystyle{IEEEtran}
\bibliography{s2lm_bib}


\begin{IEEEbiography}[{\includegraphics[width=1in,height=1.25in,clip,keepaspectratio]{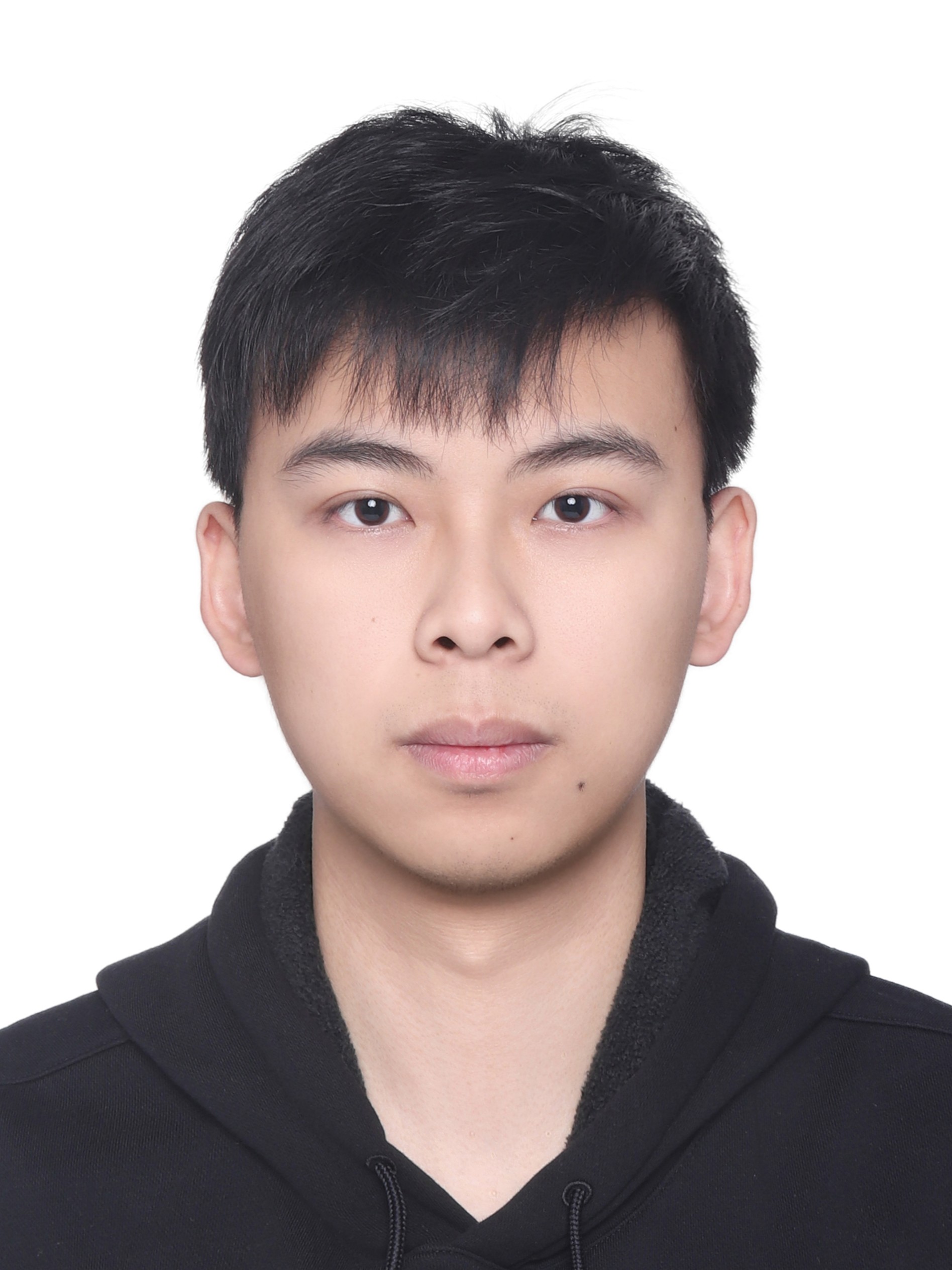}}]{Huanqi Wu}
received the B.E. degree in Computer Science and Technology from the College of Computer and Data Science, Fuzhou University, Fuzhou, China, where he is currently pursuing the Ph.D. degree. His research interests include steganography, digital watermarking, and multimedia information hiding.
\end{IEEEbiography}

\vspace{-1.5cm}

\begin{IEEEbiography}[{\includegraphics[width=1in,height=1.25in,clip,keepaspectratio]{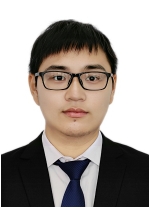}}]{Huangbiao Xu} received the B.S. degree in computer science and technology in 2021 from Fuzhou University, Fuzhou, China, where he is currently working toward the Ph.D. degree with the College of Computer and Data Science. His research interests include image and video understanding, human motion analysis, and image enhancement.
\end{IEEEbiography}

\vspace{-1.5cm}

\begin{IEEEbiography}[{\includegraphics[width=1in,height=1.25in,clip,keepaspectratio]{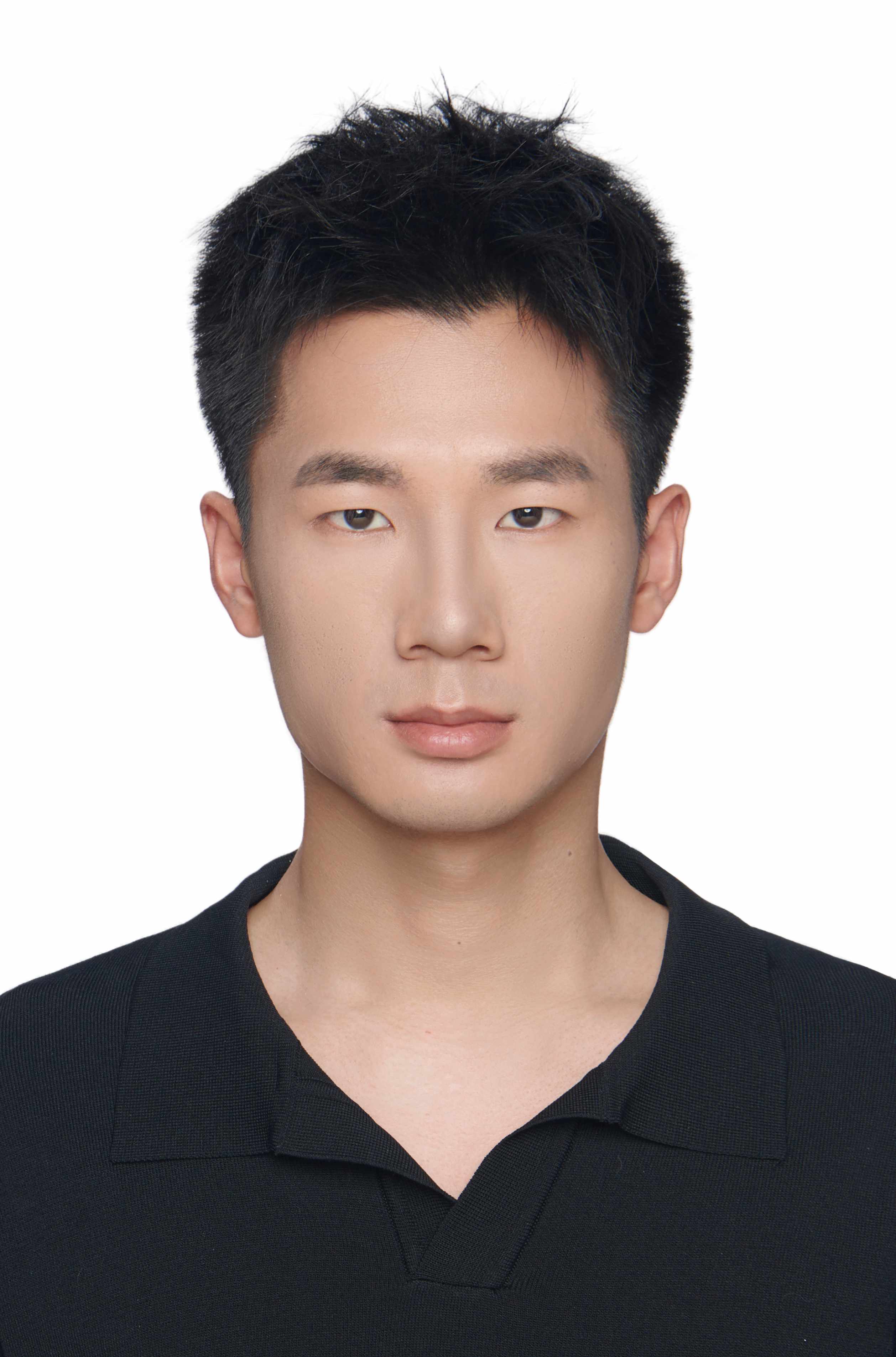}}]{Yongliang Xu} is currently a Ph.D. student with the School of Mathematics and Statistics, Fuzhou University, Fuzhou, China. From September to December 2026, he was an exchange student with the Graduate Institute of Statistics, National Central University, Taoyuan, Taiwan. His recent research interests are applied cryptography and cloud computing security.
\end{IEEEbiography}

\vspace{-1.5cm}

\begin{IEEEbiography}[{\includegraphics[width=1in,height=1.25in,clip,keepaspectratio]{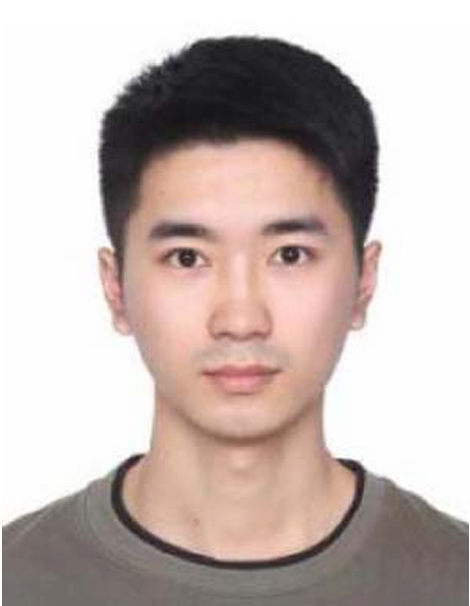}}]{Jiaxin Cai} received the BE degree from Fuzhou University. He is currently working toward the doctoral degree with the College of Computer and Data Science, Fuzhou University, China. His research interest lies primarily in computer vision.
\end{IEEEbiography}

\vspace{-1.5cm}

\begin{IEEEbiography}[{\includegraphics[width=1in,height=1.25in,clip,keepaspectratio]{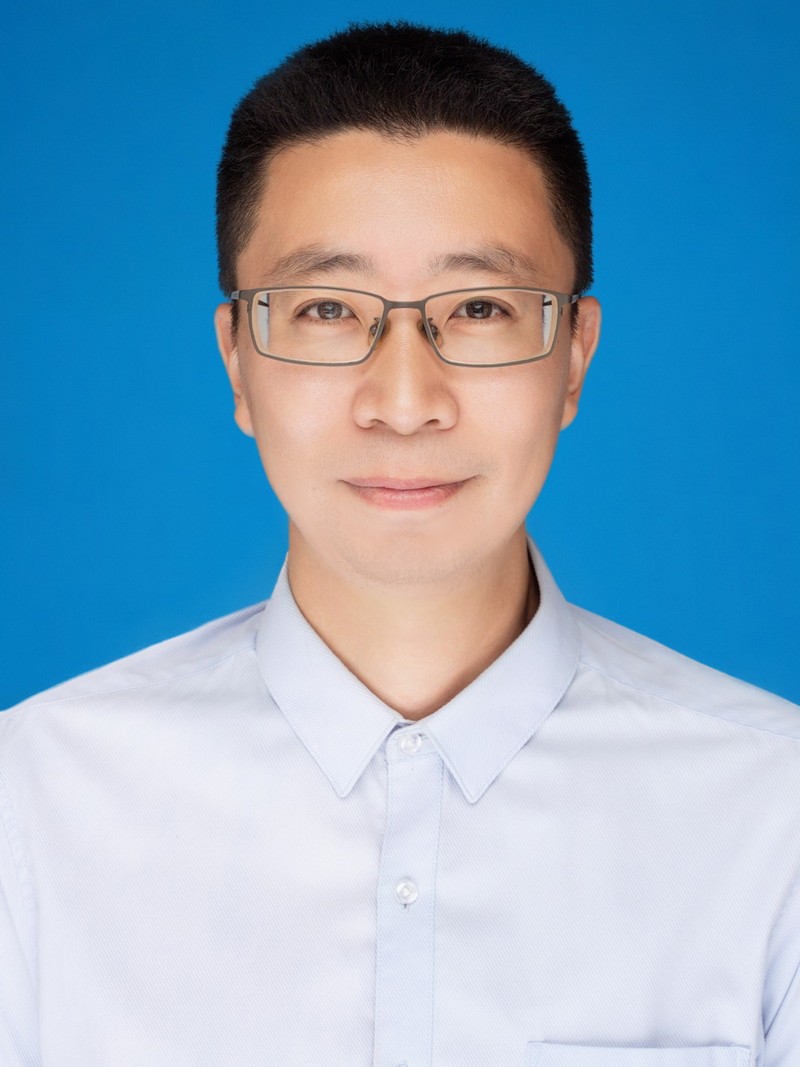}}]{Xiao Ke}
received the Ph.D. degree in Artificial Intelligence from Xiamen University, Xiamen, China, in 2011. He is currently a Full Professor with the College of Computer and Data Science at Fuzhou University. His research interests include computer vision, pattern recognition, machine learning, and artificial intelligence. He has published 100+ papers in famous journals and conferences such as: IEEE Transactions on Multimedia, IEEE Transactions on Intelligent Transportation Systems, IEEE Transactions on Circuits and Systems for Video Technology, IEEE Transactions on Image Processing, and Pattern Recognition, etc.
\end{IEEEbiography}

\vspace{-1.5cm}

\begin{IEEEbiography}[{\includegraphics[width=1in,height=1.25in,clip,keepaspectratio]{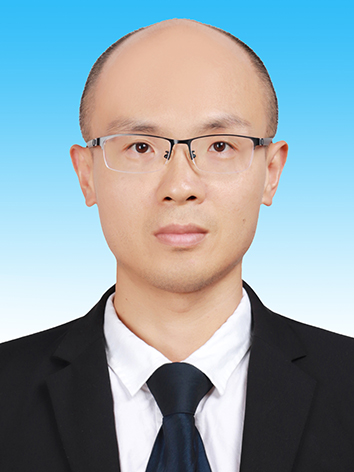}}]{Hang Cheng} (Member, IEEE) received his BS and MS degrees in applied mathematics from Fuzhou University, Fuzhou, China, in 2002 and 2005, respectively, and PhD in signal and information processing with Shanghai University, Shanghai, China, in 2016. He is a professor in the School of Mathematics and Statistics, Fuzhou University, Fuzhou, China. He is a research scholar in the School of Physical and Mathematical Sciences, Nanyang Technological University, Singapore. His current research interests include multimedia security, image processing, cryptography, and information hiding.
\end{IEEEbiography}

\end{document}